\documentclass{article}
\usepackage[accepted]{icml2021Arxiv}
\usepackage{amssymb}
\usepackage{amsmath}
\usepackage{amsthm}
\theoremstyle{plain}
\usepackage{bbm}
\usepackage{mathtools}
\usepackage{bm}
\usepackage{csvsimple}
\usepackage{dirtytalk}
\usepackage{hyperref}
\usepackage{mdframed}
\usepackage{ulem}
\newcommand{\argmin}[1]{\underset{#1}{\arg\min}\,}

\newcommand{\GER}{\textit{\texttt{MLR}}}
\newcommand{\RGER}{\textit{\texttt{R-MLR}}}
\newcommand{\SGER}{\textit{\texttt{S-MLR}}}
\newcommand{\AGER}{\textit{\texttt{A-MLR}}}
\newcommand{\BKK}{\textit{\texttt{MLR}}}
\newcommand{\BKKs}{\textit{\texttt{MLR}}}
\newcommand{\R}{\ensuremath{\mathbb{R}}}

\newcommand{\e}{\xi}
\newcommand{\be}{\beta}
\newcommand{\x}{\mathbf{X}}% changer pour grand X
\newcommand{\spa}{\gamma}
\newcommand{\LAS}{LASSO\,\,}

\newcommand{\Emu}{\mu}
\newcommand{\Sig}{\textbf{S}}
\newcommand{\bX}{\mathbb{X}}% changer en mathbb
\newcommand{\bY}{\mathbf{Y}}

\newcommand{\bv}{\mathbf{v}}
\newcommand{\cS}{\ensuremath{\mathcal{S}}}

\newcommand{\cA}{\ensuremath{\mathcal{A}}}
\newcommand{\1}{\mathbbm{1}}
\newcommand{\I}{\mathbb{I}}

\newcommand{\cD}{\ensuremath{\mathcal{D}}}

\newtheorem{mydef}{Definition}

\newtheorem{ERG-c}{\GER-Approach}

\usepackage{tikz}
\usetikzlibrary{arrows,backgrounds}
\usepgflibrary{shapes.multipart}

\begin{document}

\twocolumn[
\icmltitle{Muddling Labels for Regularization, a novel approach to generalization}

\icmlsetsymbol{equal}{*}

\begin{icmlauthorlist}
\icmlauthor{Karim Lounici}{equal,to}
\icmlauthor{Katia Meziani}{equal,goo}
\icmlauthor{Benjamin Riu}{equal,to,goo,act}
\end{icmlauthorlist}

\icmlaffiliation{to}{CMAP, Ecole Polytechnique, France}
\icmlaffiliation{goo}{University PSL Dauphine, Paris}
\icmlaffiliation{act}{Uptilab ????}

\icmlcorrespondingauthor{Benjamin Riu}{}
\icmlcorrespondingauthor{Karim Lounici}{karim.lounici@polytechnique.edu}

\icmlkeywords{Machine Learning, ICML}

\vskip 0.3in
]

\begin{abstract}
Generalization is a central problem in Machine Learning. Indeed most prediction methods require careful calibration of hyperparameters usually carried out on a hold-out \textit{validation} dataset to achieve generalization. The main goal of this paper is to introduce a novel approach to achieve generalization without any data splitting, which is based on a new risk measure which directly quantifies a model's tendency to overfit. To fully understand the intuition and advantages of this new approach, we illustrate it in the simple linear regression model ($Y=X\beta+\xi$) where we develop a new criterion. We highlight how this criterion is a good proxy for the true generalization risk. 
Next, we derive different procedures which tackle several structures simultaneously (correlation, sparsity,...). Noticeably, these procedures \textbf{concomitantly} train the model and calibrate the hyperparameters. In addition, these procedures can be implemented via 
classical gradient descent methods when the criterion is differentiable w.r.t. the hyperparameters. Our numerical experiments reveal that our procedures are computationally feasible and compare favorably to the popular approach (Ridge, \LAS and Elastic-Net combined with grid-search cross-validation) in term of generalization. They also outperform the baseline on two additional tasks: estimation and support recovery of $\beta$. Moreover, our procedures do not require any expertise for the calibration of the initial parameters which remain the same for all the datasets we experimented on.

\end{abstract}

%%%%%%%%%%%%%%%%%%%%%%%%%%%%%%%%%%%%%%%
\section*{Introduction}
%%%%%%%%%%%%%%%%%%%%%%%%%%%%%%%%%%%%%%%%%%%%

Generalization is a central problem in machine learning. 
Regularized or constrained Empirical Risk Minimization (\textit{\texttt{ERM}}) is a popular approach to achieve generalization \cite{kukavcka2017regularization}. Ridge \cite{hoerl1970ridge}, \LAS \cite{tibshirani1996regression} and Elastic-net \cite{zou2005regularization} belong to this category.
The regularization term or the constraint is added in order to achieve generalization 
and to enforce some specific structures on the constructed model (sparsity, low-rank, coefficient positiveness,...). This usually involves introducing hyperparameters which require calibration. 
The most common approach is data-splitting. Available data is partitioned into a {\it training/validation}-set. The {\it validation}-set is used to evaluate the generalization error of a model built using only the {\it training}-set.

Several hyperparameter tuning strategies were designed to perform hyperparameter calibration: Grid-search, Random search \cite{bergstra2012random} or more advanced hyperparameter optimization techniques \cite{bergstra2011algorithms, bengio2000gradient,schmidhuber1987evolutionary}. For instance, BlackBox optimization \cite{brochu2010tutorial} is used when the evaluation function is not available \cite{lacoste2014sequential}. It includes in particular Bayesian hyperparametric optimization such as Thompson sampling \cite{movckus1975bayesian,snoek2012practical,thompson1933likelihood}.
These techniques either scale exponentially with the dimension of the hyperparameter space, or requires a smooth convex optimization space \cite{shahriari2015taking}. Highly non-convex optimization problems on a high dimensionnal space can be tackled by Population based methods (Genetic Algorithms\cite{chen2018autostacker,real2017large,olson2016automating}, Particle Swarm~\cite{lorenzo2017particle,lin2008particle}) but at  a high computational cost. Another family of advanced methods, called gradient-based techniques, take advantage of gradient optimization techniques \cite{domke2012generic} like our method. They fall into two categories, Gradient Iteration and Gradient approximation. Gradient Iteration directly computes the gradient w.r.t. hyperparameters on the training/evaluation graph. 
This means differentiating a potentially lengthy optimization process which is known to be a major bottleneck \cite{pedregosa2016hyperparameter}. Gradient approximation is used to circumvent this difficulty, through implicit differentiation \cite{larsen1996design,bertrand2020implicit}. However, all these advanced methods require data-splitting to evaluate the trained model on a hold-out $validation$-set, unlike our approach.

Another approach is based on unbiased estimation of the generalization error of a model (SURE \cite{stein1981estimation}, $AIC$ \cite{akaike1974new}, $C_p$-Mallows \cite{mallows2000some}) on the $training$-set.
Meanwhile, other methods improve generalization during the training phase without using a hold-out $validation$-set. For instance, Stochastic Gradient Descent and the related batch learning techniques \cite{bottou1998online} achieve generalization by splitting the training data into a large number of subsets and compute the Empirical Risk (ER) on a different subset at each step of the gradient descent. This strategy converges to a good estimation of the generalization risk provided a large number of observations is available. Bear in mind this method and the availability of massive datasets played a crucial role in the success of Deep neural networks. Although batch size has a positive impact on generalization \cite{he2019control}, it cannot maximize generalization on its own.

Model aggregation is another popular approach to achieve generalization. It concerns for instance Random Forest \cite{ho1995random,breiman2001random}, MARS \cite{friedman1991multivariate} and Boosting \cite{freund1995desicion}. This approach aggregates weak learners previously built using bootstrapped subsets of the $training$-set. The training time of these models is considerably lengthened when a large number of weak learners is considered, which is a requirement for improved generalization. Recall XGBOOST \cite{chen2016xgboost} combines a version of batch learning and model aggregation to train weak learners.\\
MARS, Random Forest, XGBOOST and Deep learning have obtained excellent results in Kaggle competitions and other machine learning benchmarks \cite{fernandez2014we, escalera2018neurips}.
However these methods still require regularization and/or constraints in order to generalize. This implies the introduction of numerous hyperparameters which require calibration on a hold-out $validation$-set for instance \textit{via} Grid-search. Tuning these hyperparameters requires expensive human expertise
and/or computational resources.

We approach generalization from a different point of view. The underlying intuition is the following. We no longer see generalization as the ability of a model to perform well on unseen data, but rather as the ability to avoid finding pattern where none exist. Using this approach, we derive a novel criterion and several procedures which do not require data splitting to achieve generalization.

This paper is intended to be an introduction to this novel approach. Therefore, for the sake of clarity, we consider here the linear regression setting but our approach can be extended to more general settings like deep learning\footnote{In another project, we applied this approach to deep neural networks on tabular data and achieved good generalization performance. We obtained results which are equivalent or superior to Random Forest and XGBOOST. We also successfully extended our approach to classification on tabular data. This project is in the final writing phase and will be posted on Arxiv soon.}.

Let us consider the linear regression model:
\begin{eqnarray}
\label{mod2}
 \bY = \bX \be^* + \bm{\e},
\end{eqnarray}
where $\bX^\top=(\x_1,\cdots,\x_n)$ is the $n\times p$ \textit{design matrix} and  the $n$-dimensional vectors $\bY=(Y_i,\cdots,Y_n)^\top$ and $\bm{\e}=(\e_1,\cdots,\e_n)^\top$ are respectively the response and the noise variables. Throughout this paper, the noise level $\sigma>0$ is unknown. 
Set $||\bv||_n=(\frac{1}{n}\sum_{i=1}^n v_i^2)^{1/2}$ for any $\bv=(v_1,\ldots,v_n)^\top\in \R^n$. 

In practice, the correlation between $\x_i$ and $Y_i$ is unknown and may actually be very weak. In this case, $\x_i$ provides very little information about $Y_i$ and we expect from a good procedure to avoid building a spurious connection between $\x_i$ and $Y_i$. Therefore, by understanding generalization as \say{\textit{do not fit the data in non-informative cases}}, we suggest creating an artificial dataset which preserves the marginal distributions while the link between $\x_i$ and $Y_i$ has been completely removed. A simple way to do so is to construct an artificial set $\widetilde{\cD}=({\mathbb{X}},\widetilde{\mathbf{Y}}) = (\bX,\pi(\mathbf{Y}))$ by applying permutations $\pi \in \mathfrak{S}_n$ (the set of permutations of $n$ points) on the components of $\bY$ of the initial dataset $\cD$ where for any $\mathbf{y}\in \mathbb{R}^n$, we set $\pi(\mathbf{y}) = (y_{\pi(1)},\ldots,y_{\pi(n)})^\top$.\\

The rest of the paper is organized as follows. In Section \ref{Sec:GER} we introduce our novel criterion and highlight its generalization performance. %in the regression setting. 
In Section \ref{Sec:proc}, this new approach is applied to several specific data structures in order to design \textbf{adapted} procedures which are compatible with gradient-based optimization methods. We also point out several advantageous points about this new framework in an extensive numerical study.
Finally we discuss possible directions for future work in Section \ref{sec:con}.

%%%%%%%%%%%%%%%%%%%%%%%%%%%%%%%%%%%%%%%%
\section{Label muddling criterion}
\label{Sec:GER}
%%%%%%%%%%%%%%%%%%%%%%%%%%%%%%%%%%%%%%%%%%

In model \eqref{mod2}, we want to recover $\beta^*$ from $\mathcal{D} = (\bX,\bY)$. Most often, the  $n$ observations are partitioned into two parts of respective sizes  $n_{train}$ and $n_{val}$, which we denote the $train$-set $(\bX_{train},\bY_{train})$ and the $validation$-set $(\bX_{val},\bY_{val})$ respectively. The $train$-set is used to build a family of estimators $\{\beta(\theta, \bX_{train}, \bY_{train})\}_{\theta}$ which depends on a hyperparameter $\theta$. Next, in order to achieve generalization, we use the $validation$-set to calibrate $\theta$. This is carried out by minimizing the following empirical criterion w.r.t. $\theta$:
\begin{equation*}
\|\bY_{val} - \bX_{val} \,\beta(\theta, \bX_{train}, \bY_{train}) \|_{n_{val}}.
\end{equation*}
In our approach, we use the complete dataset $\mathcal{D}$ to build the family of estimators and to calibrate the hyperparameter $\theta$.  

\begin{mydef}
Fix $T \in \mathbb{N}^*$. Let $\{\pi^t\}_{t=1}^T$ be $T$ permutations in $\mathfrak{S}_n$. Let $\{\beta(\theta,\cdot,\cdot)\}_{\theta}$ be a family of estimator.

The \textbf{\BKKs\, criterion\footnote{Muddling Labels for Regularization}} is defined as:
\begin{small}\begin{align}
\label{eq:cBKKs}
&\BKK_{\beta}(\theta) = \Vert \bY - \bX \be(\theta, \bX, \bY)\Vert_n\notag\\
&\hspace{2cm}- \frac{1}{T} \sum_{t=1}^{T} \Vert \pi^t(\bY) - \bX\be(\theta, \bX, \pi^t(\bY)) \Vert_n.
\end{align}
\end{small}
\end{mydef}

The \GER\, criterion performs a trade-off between two antagonistic terms. The first term fits the data while the second term prevents overfitting. Since the \GER\, criterion is evaluated directly on the whole sample without any hold-out $validation$-set, this approach is particularly useful for small sample sizes where data-splitting approaches can produce strongly biased performance estimates \cite{vabalas2019,varoquaux2018}.\\
We highlight in the following numerical experiment the remarkable generalization performance of the \GER\, criterion.

\paragraph{NUMERICAL EXPERIMENTS.}
\noindent\\
%%%%%%%%%%%%%%%%%%%%%%%%%%%%%%%%%%%%%%%%%%%

We consider two families 
$$\mathcal F(\theta)=\{ \be(\theta,\mathbb{X}_{train},\mathbf{Y}_{train})  \}_{\theta}
$$
of estimators constructed on a $train$-dataset and indexed by $\theta$: Ridge and \LAS. We are interested in the problem of calibration of the hyperparameter $\theta$ on a grid $\Theta$.
We compare two criteria $\mathcal C(\theta)$ to calibrate $\theta$: the \GER\, criterion and  cross-validation (implemented as Ridge, Lasso in Scikit-learn \cite{pedregosa2011scikit}). For each criterion $\mathcal C(\theta)$, the final estimator $\widehat{\be}_{train}=\be(\widehat\theta,\mathbb{X}_{train},\mathbf{Y}_{train})$ is $s.t.$
$$\widehat\theta=\underset{\theta\in \Theta}{\arg\min}\,\mathcal C(\theta).
$$
%
%%%%%%%%%%%%%%%%%%%%%%%%%%%%%%%%%%%%%%%%%%%
\paragraph{$\mathbf{R^2}$-score.}
For each family, the generalization performance of each  criterion is evaluated using the hold-out $test$-dataset $\mathcal D_{test}=(\bX_{test},\bY_{test})$ by computing the following $\mathbf{R^2}$-scores:
\begin{eqnarray}
\label{R2}
\mathbf{R^2}(\widehat{\be}_{train})=1- \frac{\Vert \bY_{test} - \bX_{test}\,\widehat{\be}_{train} \Vert_2}{\Vert \bY_{test} -\overline{\bY}_{test}\1_n \Vert_2}\,(\leq 1),
\end{eqnarray}
where $\overline{\bY}_{test}$ is the empirical mean of $\bY_{test}$. For the sake of simplicity, we set $\mathbf{R^2}(\widehat\theta)=\mathbf{R^2}(\widehat\be_{train})$. The oracle we aim to match, is
$$
\theta^*=\underset{\theta\in \Theta}{\arg\min}\,\mathbf{R^2}(\theta).
$$
Our first numerical experiments concern synthetic data\footnote{Our Python code is released as an open source package for replication in the github repository \href{https://github.com/AnonymousICML2021Submitter/ICML2021SupplementaryMaterial}{ICML2021SupplementaryMaterial}.}.

%%%%%%%%%%%%%%%%%%%%%%%%%%%%%%%%%%%%%%%%%%
\paragraph{Synthetic data.} For $p=80$, we generate observations $(\x,Y) \in \R^p\times \R$, $s.t.$ 
$Y = \x^\top \beta^* + \epsilon$, with  $\epsilon\sim\mathcal N(0,\sigma)$, $\sigma=10$ or $50$. We consider three different scenarii. \textbf{Scenario A} (correlated features) corresponds to the case where the \LAS is prone to fail and Ridge should perform better. \textbf{Scenario B} (sparse setting) corresponds to a case known as favorable to \LAS. \textbf{Scenario C} combines sparsity and correlated features. 
For each scenario we sample a $train$-dataset of size $n_{train}=100$ and a $test$-dataset of size $n_{test}=1000$.\\
For each scenario, we perform $M=100$ repetitions of the data generation process to produce $M$ pairs of $train$/$test$ datasets. Details on the data generation process can be found in the Appendix.

%%%%%%%%%%%%%%%%%%%%%%%%%%%%%%%%%%%%%%%%%%%
\paragraph{Performances evaluation.}
For each family $\mathcal F(\theta)$ and each criterion $\mathcal C(\theta)$, we construct on every $train$-dataset, the corresponding model $\widehat{\be}_{train}$. Next, using the corresponding hold-out $test$-dataset $\mathcal D_{test}=(\bX_{test},\bY_{test})$, we compute their $\mathbf{R^2}$-scores.

%
%%%%%%%%%%%%%%%%%%%%%%%%%%%%%%%%%%%%%%%%%%%
\paragraph{Impact of $\,T$.} 

\begin{figure}[http]
\centering
\includegraphics[scale=0.30]{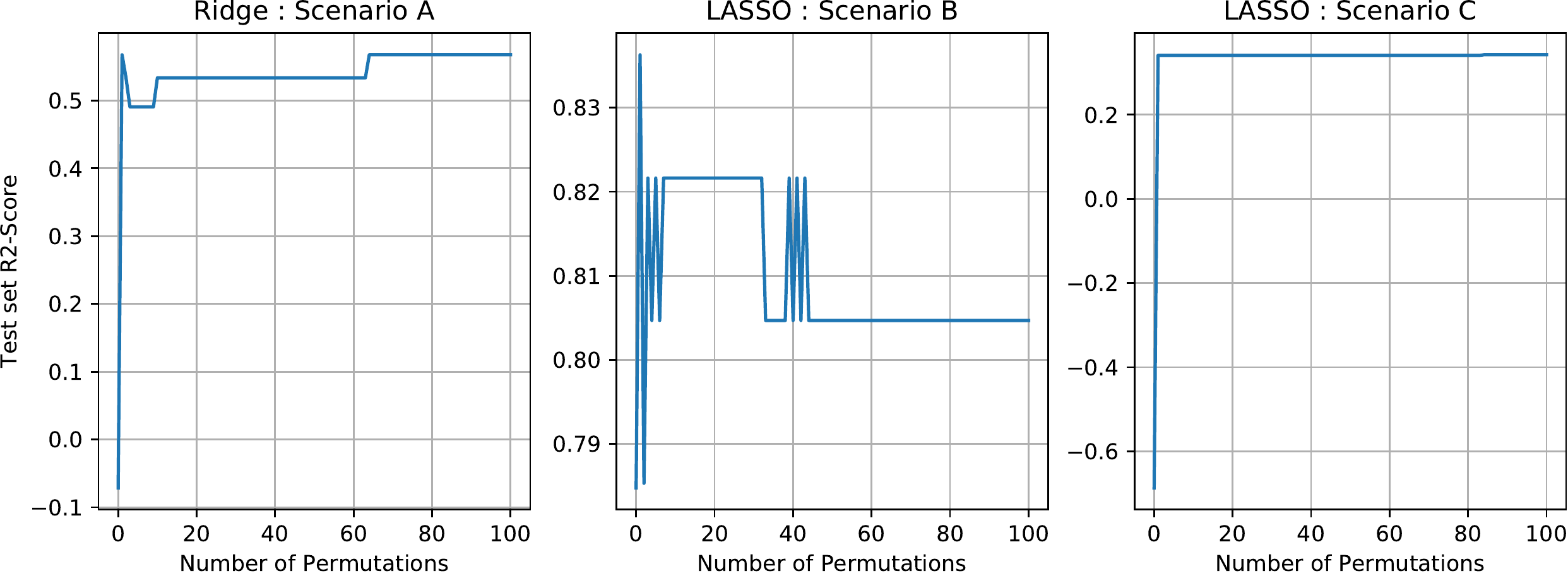}
\label{fig:1predK}
\caption{Variation of the $R^2$-score w.r.t. the number of permutation.}
\end{figure}

In Figure~\ref{fig:1predK} we study the impact of the number of permutations $T$ on the generalization performance of the criterion measured via the $R^2$-score in \eqref{R2}. The most striking finding is the sharp increase in the generalization performance from the first added permutation in \textbf{Scenarios A and C}. Adding more permutations does not impact the generalization but actually improves the running time and stability of the novel procedures which we will introduce in the next section. In a pure sparsity setting (\textbf{Scenario B} with LASSO), adding permutations marginally increases the generalization.

\begin{figure}[http]
\centering
\includegraphics[scale=0.31]{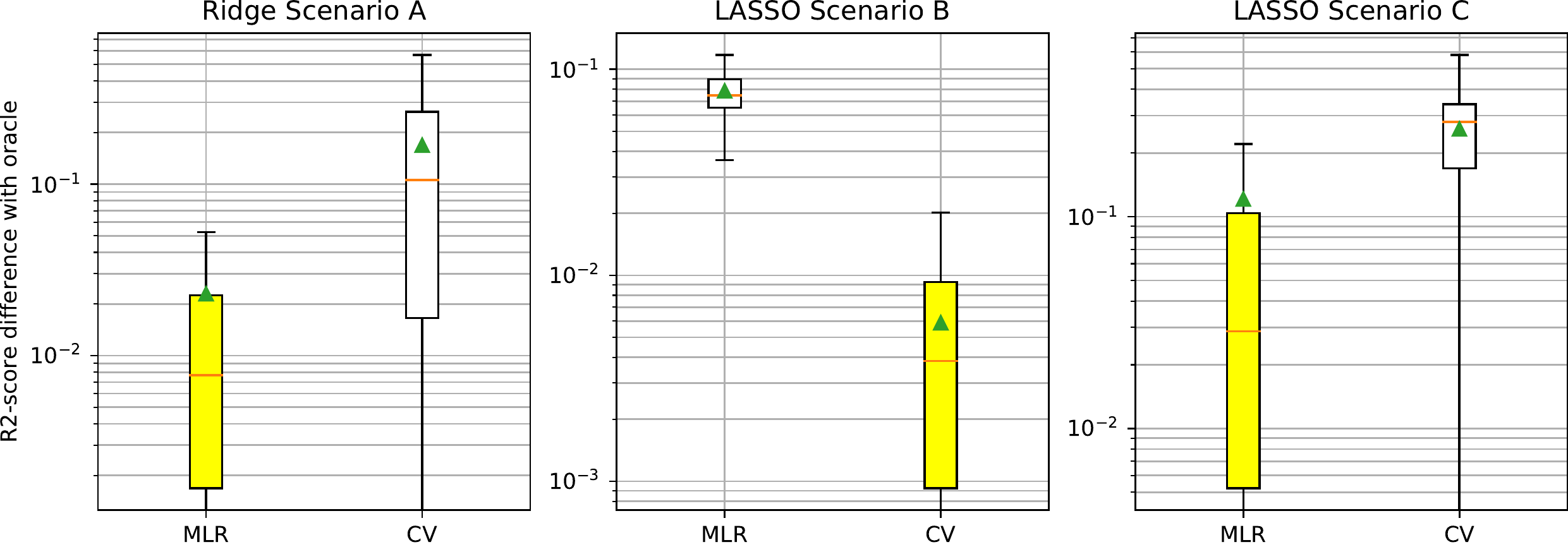}
\label{fig:2pred}
\caption{Generalization performance for \BKK and CV for Ridge and \LAS Hyperparameter calibration}
\end{figure}

%
%%%%%%%%%%%%%%%%%%%%%%%%%%%%%%%%%%%%%%%%%%%
\paragraph{Comparison of generalization performance.} 
For \textbf{Scenario A} with the Ridge family, we compute the difference
$\mathbf{R^2}(\widehat\theta^{\mathcal C(\theta)}) -  \mathbf{R^2}(\theta^{*}),
$
for the two criteria $\mathcal C(\theta)$ (\BKK\, and CV).
For \textbf{Scenarios B and C}, we consider the \LAS family and compute the same difference. Boxplots in Figure~\ref{fig:2pred} summarize our finding over 100 repetitions of the synthetic data. The empirical mean is depicted by a green triangle on each boxplot. 
Moreover, to check for statistically significant margin in $\mathbf{R^2}$-scores between different procedures, we use the Mann-Whitney test (as detailed in \cite{kruskal1957historical} and implemented in scipy \cite{virtanen2020scipy}). The boxplots highlighted in yellow correspond to the best procedures according to the Mann-Whitney (MW) test.
As we observed, the \BKK\, criterion performs better than CV for the calibration of the Ridge and LASSO hyperparameters in \textbf{Scenarios A and C} in the presence of correlation in the design matrix.

%%%%%%%%%%%%%%%%%%%%%%%%%%%%%%%%%%%%%%%%%%%%%%%%%%%%%%
\paragraph{Plot of the generalization performance.}

In order to plot the different criteria together, we use the following rescaling. Let $F\,:\, \Theta \rightarrow \R$, we set $$
\underline{\theta}=\underset{\theta\in \Theta}{\arg\min}\,F(\theta) \,\text{ and }\,\overline{\theta}=\underset{\theta\in \Theta}{\arg\max}\,F(\theta). 
$$
For any $\theta\in \Theta$, we define 
$$
\Lambda_{F}(\theta) = \frac{F(\theta) - F(\underline{\theta})}{F(\overline{\theta}) - F(\underline{\theta})}.
$$

\begin{figure}[htp]
\centering
\includegraphics[scale=0.30]{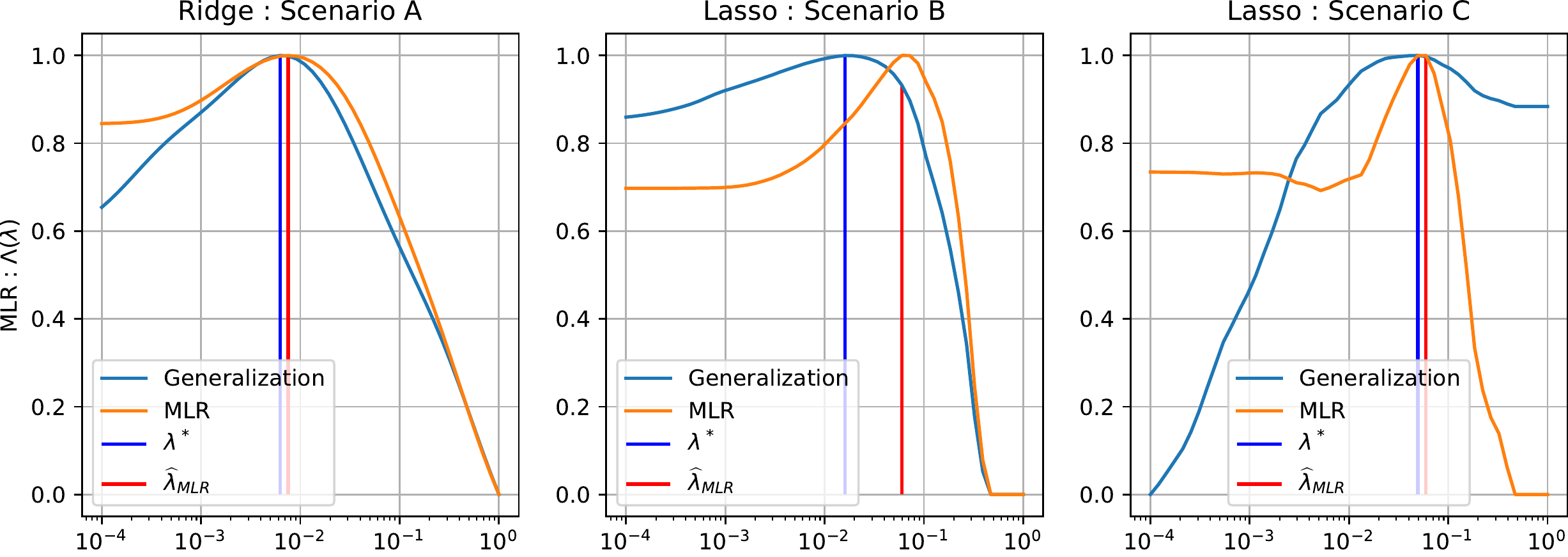}
\includegraphics[scale=0.30]{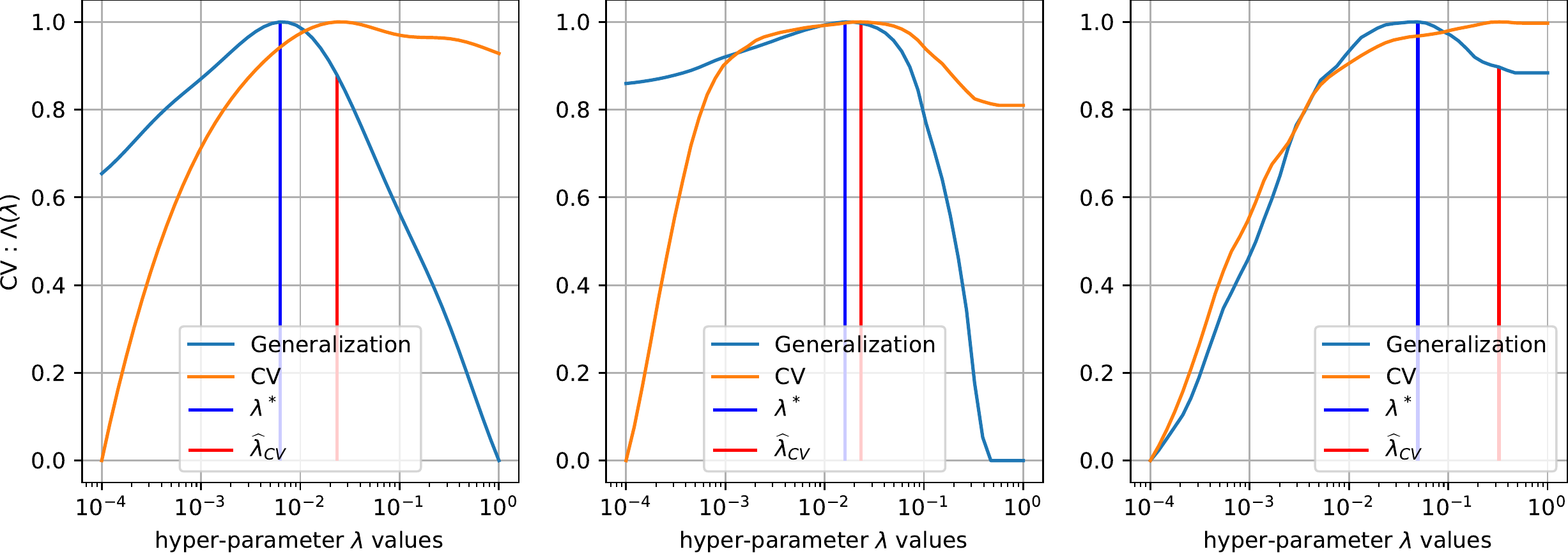}
\caption{Criterion landscape for grid-search calibration with \BKK\, (\textbf{Up}) and CV (\textbf{down}). 
}
\label{fig:3pred}
\end{figure}

Figure \ref{fig:3pred} contains the rescaled versions of the $\mathbf{R}^2$-score on the test set and the $\BKK$ and $CV$ criteria computed on the $train$-dataset. The vertical lines correspond to the selected values of the hyperparameter in the grid $\Theta$ by the \BKK\, and CV criteria as well as the optimal hyperparameter $\theta^*$ for the $\mathbf{R}^2$-score on the test set. The \BKK\, criterion is a good proxy for the generalization performance ($R^2$- score on the $test$-dataset) on the whole grid $\Theta$ for the Ridge in \textbf{Scenario A}.
In \textbf{Scenario C} with the LASSO, 
the \BKK\, criterion is a smooth function with steep variations in a neighborhood of its global optimum. This is an ideal configuration for the implementation of gradient descent schemes.

the \BKK\,criterion performs better than CV on \textbf{correlated} data for grid-search calibration of the LASSO and Ridge hyperparameters. However CV works better in the pure sparsity scenario. This motivated the introduction of novel procedures based on the \BKK\, principle which can handle the sparse setting better.

%%%%%%%%%%%%%%%%%%%%%%%%%%%%%%%%%%%%%%%%%
\section{Novel procedures}
\label{Sec:proc}
%%%%%%%%%%%%%%%%%%%%%%%%%%%%%%%%%%%%%%%%%

In Section \ref{Sec:GER}, we used (\ref{eq:cBKKs}) to perform grid-search calibration of the hyperparameter. However, if $\{\beta(\theta)\}_{\theta}$ is a family of models differentiable w.r.t. $\theta$, we can minimize (\ref{eq:cBKKs}) w.r.t. $\theta$ via standard gradient-based methods. This motivated the introduction of new procedures based on the \GER\, criterion.

\begin{mydef}
\label{def:bkkproc}
Consider $\{\beta(\theta)\}_{\theta}$ a family of models differentiable  w.r.t. $\theta$. 
%Fix $T \in \mathbb{N}^*$. 
Let $\{\pi^t\}_{t=1}^T$ be $T$ derangements in $\mathfrak{S}_n$. The $\beta$-\textbf{\GER\, procedure} is
\begin{eqnarray}\label{eq:pBKK1}
\begin{array}{lcl}
  \widehat \be =\be(\widehat\theta)&\text{with} 
  &\widehat{\theta} = \argmin{\theta} \GER_{\be} (\theta).
\end{array}
\end{eqnarray}
where $\GER_\beta$ is defined in \eqref{eq:cBKKs}.
\end{mydef}

Moreover, using our approach, we can also enforce several \textbf{additional structures} simultaneously (sparsity, correlation, group sparsity, low-rank,...) by constructing appropriate families of models. In this regard, let us consider the 3 following procedures which do not require a hold-out $validation$-set. 

\paragraph{\RGER\, procedure for correlated designs.}
%%%%%%%%%%%%%%%%%%%%%%%%%%%%%%%%%%%%%%%%%%%%%%%%%%%%%%%%%

The Ridge family of estimators $\{\be^R(\lambda, \bX, \bY)\}_{\lambda>0}$ is defined as follows:
\begin{align}
\label{Ridge}
\be^R(\lambda, \bX, \bY) = (\bX^\top \bX + \lambda \I_p)^{-1}\bX^\top \bY,\quad \lambda>0.
\end{align}

Applying Definition \ref{def:bkkproc} with the Ridge family, we obtain the \RGER\, procedure
$
\widehat{\beta}^{\mathcal{R}} = \beta^{R}(\widehat{\lambda})
$
where $\widehat{\lambda} = \argmin{\lambda>0}
\GER_{\beta^{R}}(\lambda)$. This new optimisation problem can be solved by gradient descent, contrarily to the previous section where we performed a grid-search calibration of $\lambda$.

\paragraph{\SGER\, for sparse models.}
%%%%%%%%%%%%%%%%%%%%%%%%%%%%%%%%%%%%%%%%%
We design in Definition \ref{sparsefunction} below  a novel differentiable family of models to enforce sparsity in the trained model. Applying Definition \ref{def:bkkproc} to this family, we can derive the \SGER\, procedure: $\widehat{\beta}^{\cS} = \beta^{\cS}(\widehat{\theta})$ 
where $\widehat{\theta} = \argmin{\theta}\, \GER_{\be^\cS}(\theta)$. 

\begin{mydef}
\label{sparsefunction} 
 Let $\{\beta^{\cS}(\lambda,\kappa,\spa,\bX,\bY)\}_{(\lambda,\kappa,\spa)\in\R^*_+\times\R^*_+\times\R^p}\,\,$ be a family closed-form estimators defined as follows:
\begin{align}
\label{Sbeta}
\be^{\cS}(\lambda,\kappa,\spa, \bX, \bY) = \cS(\kappa,\spa) \beta^R(\lambda, \bX \cS(\kappa,\spa), \bY),
\end{align}
where $\be^R$ is defined in \eqref{Ridge}, the \textbf{\textit{quasi-sparsifying}} function $\cS:\R^*_+ \times\R^p \rightarrow ]0,1[^{p\times p}$ is $s.t.$
$$\cS(\kappa,\spa)=\text{diag}\left(\cS_1(\kappa,\spa),\cdots,\cS_p(\kappa,\spa)\right),$$
where for any $j=1,\cdots,p$,
$$
\begin{array}{lll}
\cS_j \ : &\R^*_+ \times\R^p \rightarrow ]0,1[ \\
&(\kappa,\spa) \mapsto\cS_j (\kappa,\spa)=\left(1+e^{-\kappa \times (\sigma_\gamma^2 + 10^{-2})(\spa_j - \overline \spa)}\right)^{-1},
\end{array}
$$
with $\overline \spa=\frac{1}{p}\sum_{i=1}^p \spa_i$ and $\sigma_\gamma^2=\sum_{i=1}^p (\spa_i-\overline \spa)^2$.
%$$
\end{mydef}
%

%%%%%%%%%%%%%%%%%%%%%%%%%%%%%%%%%%%%%%%%
The new family \eqref{Sbeta} enforces sparsity on the regression vector but also directly onto the design matrix. Hence it can be seen as a combination of data-preprocessing (performing feature selection) and model training (using the ridge estimator).\\
Noticeably, the \say{\textit{quasi-sparsifying}} trick transforms feature selection (a discrete optimization problem) into a continuous optimization problem which is solvable via classical gradient-based methods.
The function $\cS$ produces diagonal matrices with diagonal coefficients in $]0,1[$. Although the sigmoid function $\cS_j$ cannot take values 0 or 1, for very small or large values of $\gamma_j$, the value of the corresponding diagonal coefficient of $S(\kappa,\gamma)$ is extremely close to $0$ or $1$. In those cases, the resulting model is weakly sparse in our numerical experiments. Thresholding can then be used to perform feature selection.

%%%%%%%%%%%%%%%%%%%%%%%%%%%

\paragraph{\AGER\, for correlated designs and sparsity.}
Aggregation is a statistical technique which combines several estimators in order to attain higher generalization performance \cite{tsybakov2003}. We propose in Definition~\ref{ABKK} a new aggregation procedure to combine the estimators \eqref{Ridge} and \eqref{Sbeta}.
This essentially consists in an interpolation between $\beta^R$ and $\beta^\cS$ models, where the coefficient of interpolation is quantified via the introduction of a new regularization parameter $\mu \in \R$.

\begin{mydef}
\label{ABKK}
%\label{Elasticspecificity}
 We consider the family of models
 $$\{\beta^{\cA}(\lambda,\kappa,\spa,\mu, \bX, \bY)\}_{(\lambda,\kappa,\spa,\mu)\in\R^*_+\times\R^*_+\times\R^p\times\R}\,\,
 $$ with
\begin{align}
\label{Abeta}
\be^{\cA}(\theta, \bX,\bY) &=  \Sig(\Emu) \times \beta^R(\lambda, \bX,\bY) \notag\\
&\hspace{1cm}+ (1 - \Sig(\Emu)) \times \be^{\cS}(\lambda, \kappa,\spa, \bX,\bY),
\end{align}
where $\be^R(\lambda,\bY)$ and $\be^{\cS}(\lambda,\kappa, \spa, \bX, \bY)$ are defined in \eqref{Ridge} and \eqref{Sbeta} respectively and $\Sig$ is the sigmoid\footnote{For $\Emu\in\R$,  $\Sig(\Emu)$ takes values in $(0,1)$ and is actually observed in practice to be close to $0$ or $1$.}.
\end{mydef}

%%%%%%%%%%%%%%%%%%%%%%%%%%%

Applying Definition \ref{def:bkkproc} to this family, we can derive the \AGER\, procedure: $\widehat{\beta}^{\cA} = \beta^{\cA}(\widehat{\theta})$ 
where $\widehat{\theta} = \argmin{\theta}\, \GER_{\be^\cA}(\theta)$. 

This procedure is designed to handle both correlation and sparsity.

\begin{figure}[http!]
\centering
\includegraphics[scale=0.32]{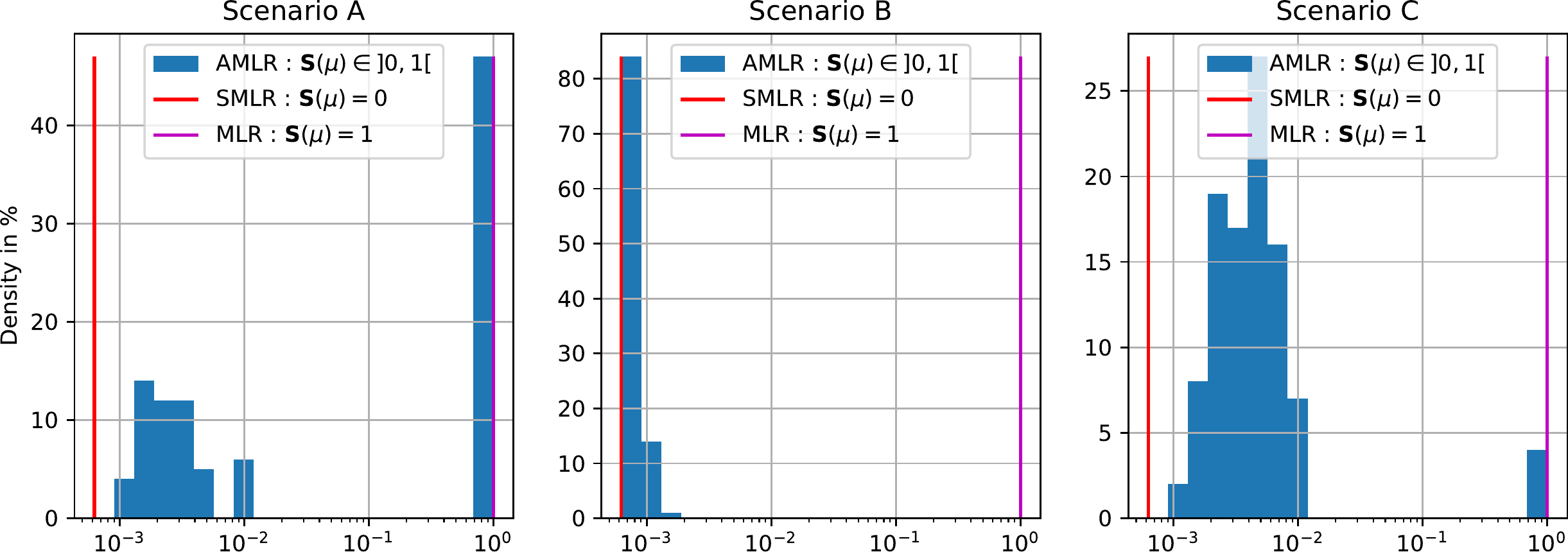}
\caption{Distribution of the value of $\textbf{S}(\widehat{\mu})$ over $100$ repetitions on synthetic data.} 
\label{fig:aggregation}
\end{figure}

Figure \ref{fig:aggregation} shows $\widehat{\beta}^{\cA}$ behaves almost as a selector which picks the most appropriate family of models between $\beta^{R}$ and $\beta^{\cS}$. Indeed, in all scenarios, $\textbf{S}(\widehat{\mu})$ only takes values close either to $0$ or $1$ in order to adapt to the structure of the model.
Moreover in \textbf{Scenario B} (sparsity), $\widehat{\be}^{\cA}$ always selects the sparse model. Indeed we always have $\textbf{S}(\widehat{\mu}) \leq 0.002$ over $100$ repetitions.

\paragraph{Algorithmic complexity.}
%%%%%%%%%%%%%%%%%%%%%%%%%%%%%%%%%%%%%%%%%%

%
Using the \BKKs\, criterion, we develop fully automatic procedures to tune regularization parameters while simultaneously training the model in a single run of the gradient descent algorithm without a hold-out $validation$ set. The computational complexity of our methods is $O(n(p+r)K)$ where $n,p,r,K$ denote respectively the number of observations, features, regularization parameters and iterations of the gradient descent algorithm. The computational complexity of our method grows only arithmetically w.r.t. the number of regularization parameters.

\paragraph{NUMERICAL EXPERIMENTS.}
%\noindent\\
%%%%%%
We performed numerical experiments on the synthetic data from Section \ref{Sec:GER} and also on real datasets described below.

%%%%%%%%%%%%%%%%%%%%%%%%%%%%%%%%%%%%%%%%%%%
\paragraph{Real data.}
We test our methods on several commonly used real datasets (UCI \cite{asuncion2007uci} and Svmlib \cite{chang2011libsvm} repositories). See Appendix for more details. Each selected UCI dataset is splitted randomly into a $80\%$ $train$-dataset and a $20\%$ $test$-dataset. We repeat this operation $M=100$ times to produce $M$ pairs of $(train,test)$-datasets.\\
In order to test our procedures in the setting $n \leq p$, we selected, from Svmlib, the news20 dataset which contains a $train$ and a $test$ dataset. We fixed the number of features $p$ and we sample six new 20news $train$-datasets  of different sizes $n$ from the initial news20 $train$-dataset. For each size $n$ of dataset, we perform $M=100$ repetitions of the sampling process to produce $M$ $train$-datasets. We kept the initial $test$-set for the evaluation of the generalization performances.

\paragraph{Number of iterations.}
%%%%%%%%%%%%%%%%%%%%%%%%%%%%%%%%%%%%%%%%%%

% Iterations

\begin{figure}[htp]
\centering
\includegraphics[scale=0.32]{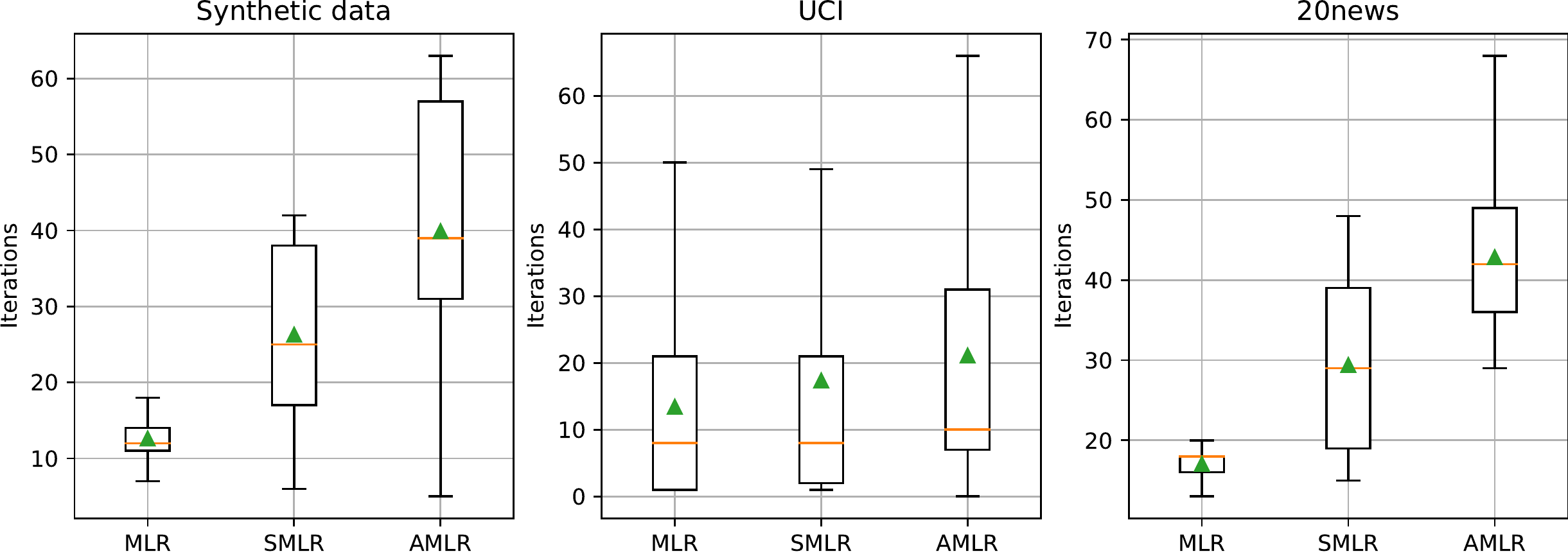}
\caption{\textbf{Synthetic, UCI and 20news data}:  Number of iterations}
\label{fig:itergraph}
\end{figure}

We choose to solve (\ref{eq:pBKK1}) using ADAM but other GD methods could be used. Figure \ref{fig:itergraph} contains the boxplots of the number of ADAM iterations for the \BKKs\, procedures on the synthetic and real datasets over the $M=100$ repetitions. Although $\BKKs_{\beta^{\cS}}$ and $\BKKs_{\beta^{\cA}}$ are highly non-convex, the number of iterations required for convergence is always about a few several dozen in our experiments. This was already observed in other non-convex settings \cite{kingma2014adam}.

\paragraph{Running time.}
%%%%%%%%%%%%%%%%%%%%%%%%%%%%%%%%%%%%%%%%%%

Our procedures were coded in Pytorch to underline how they can be parallelized on a GPU. A comparison of the running time with the benchmark procedures is not pertinent as they are implemented on cpu by Scikit-learn. The main point of our experiments was rather to show how the \GER\, procedures can be successfully parallelized. This opens promising prospects for the \GER\, approach in deep learning frameworks.

\begin{figure}[http]
    \centering
    \begin{minipage}{0.23\textwidth}
        \centering
        \includegraphics[width=0.9\textwidth]{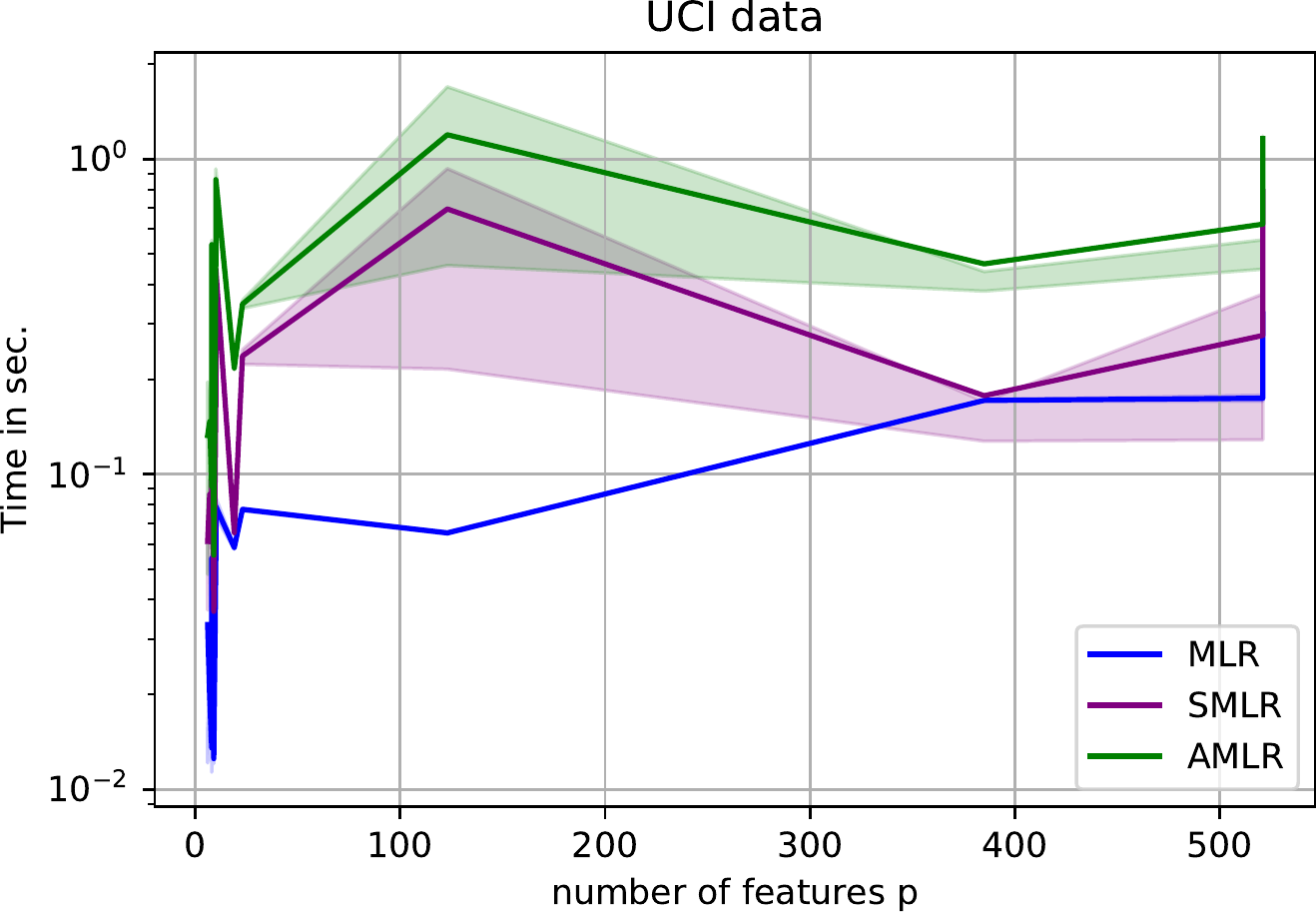} 
        \caption{\textbf{UCI data:} running time as a function of $n$}
        \label{figA:graphtime1}
    \end{minipage}\hfill
    \begin{minipage}{0.23\textwidth}
        \centering
        \includegraphics[width=0.9\textwidth]{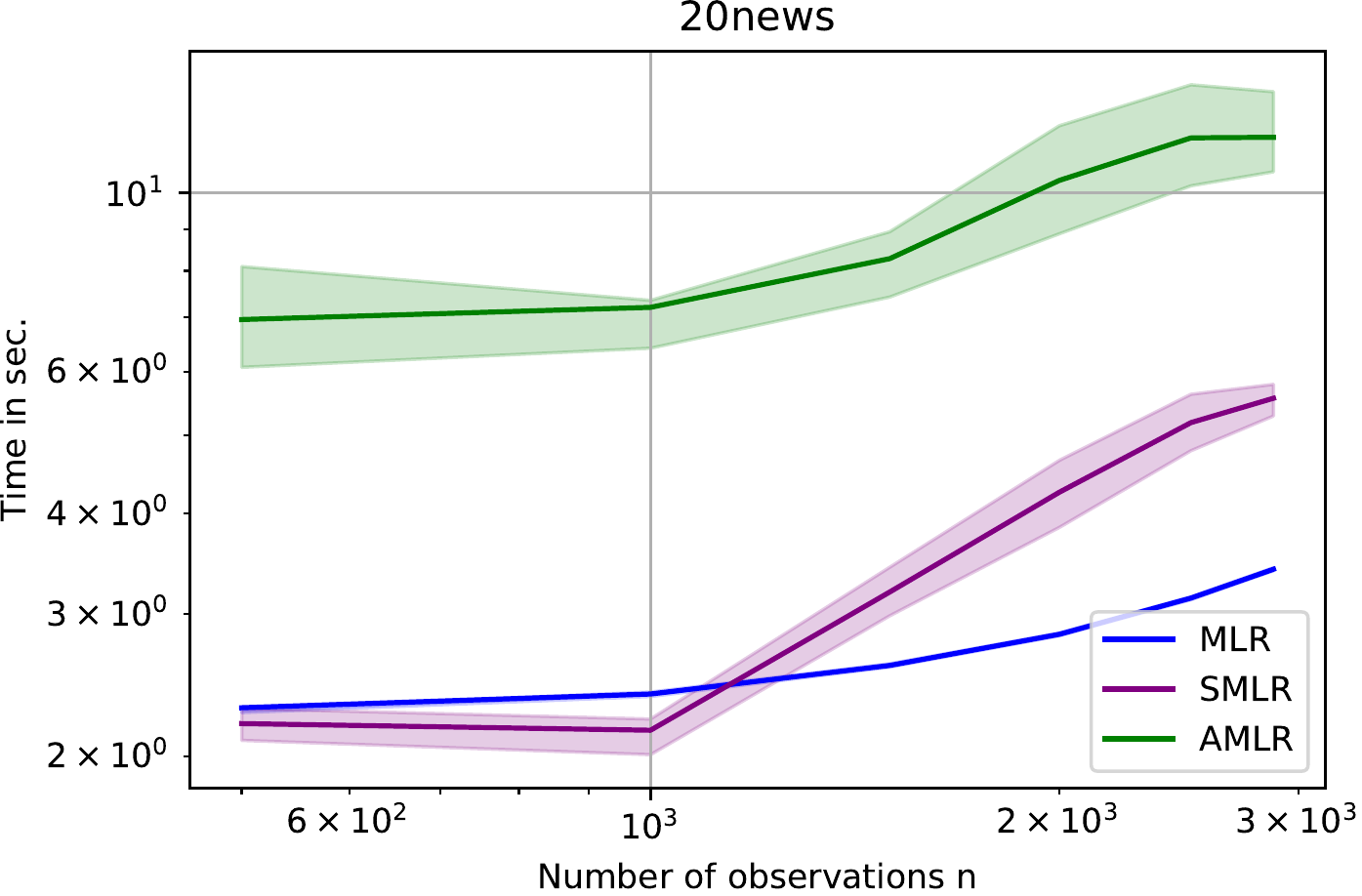}
        \caption{\textbf{20news data:} running times as a function of $p$}
        \label{figA:graphtime2}
    \end{minipage}
\end{figure}

From a computational point of view, the matrix inversion in \eqref{Ridge} is not expensive in our setting as long as the covariance matrix can fully fit on the GPU\footnote{Inversion of a $p\times p$ matrix has a $p^2$ complexity on CPU, but parallelization schemes provide linear complexity on GPU when some memory constraints are met \cite{Murugesan2018Embarrassingly,rajibr2010accelerating,chrzeszczyk2013matrix}.}. Figures \ref{figA:graphtime1} and \ref{figA:graphtime2} confirm the running time is linear in $n,p$ for the \BKKs\, procedures. This confirms the \BKK\, procedures are scalable.

\begin{figure}[http!]
\centering
 \includegraphics[scale=0.32]{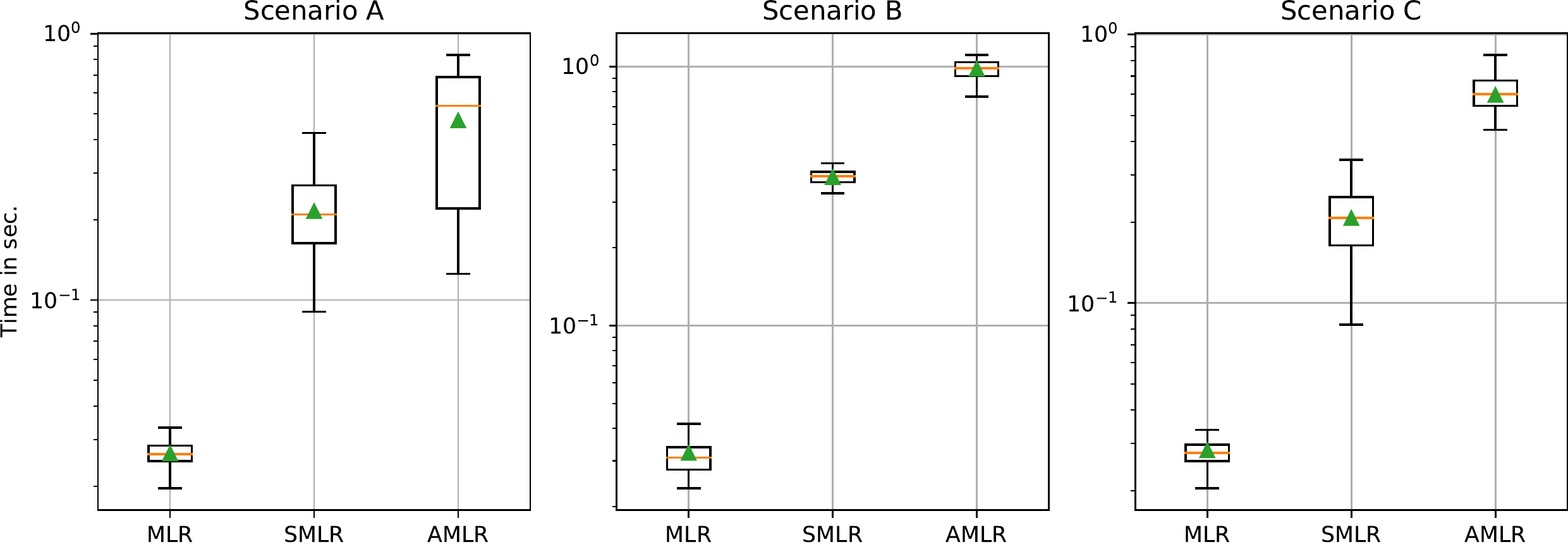}
\caption{\textbf{Synthetic data:} running times in seconds.}
\label{fig:syntheticTime}
 \end{figure}

\begin{figure}[http]
\centering
\includegraphics[scale=0.32]{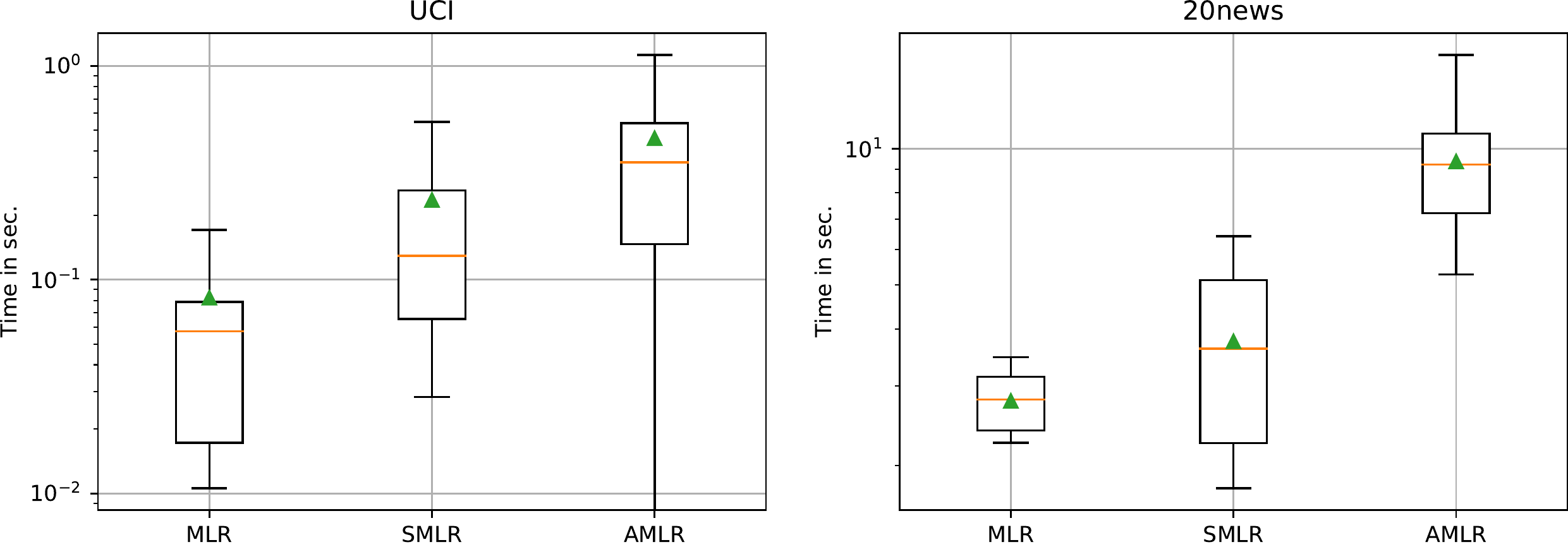}
\caption{\textbf{UCI data} (left) and \textbf{20news data} (right): running times %in seconds
.}
\label{fig:graphtime}
\end{figure}

Consequently, our procedures run in reasonable time as illustrated in Figures~\ref{fig:syntheticTime}  and \ref{fig:graphtime}.


\paragraph{Initial parameters.}
%%%%%%%%%%%%%%%%%%%%%%%%%%%%%%%%%%%%%%%%%%

Strinkingly, the initial values of the parameters (see Table~\ref{Tab:parameter}) used to implement our \BKKs\, procedures could remain the same for all the datasets we considered while still yielding consistently good prediction performances. These initial values were calibrated only once in the standard setting ($n\geq p$) on the Boston dataset \cite{harrison1978hedonic,belsley2005regression} which we did not include in our benchmark when we evaluated the performance of our procedures. We emphasize again we used these values without any modification on all the synthetic and real datasets. The synthetic and UCI datasets fall into the standard setting. Meanwhile, the 20news datasets correspond to the high-dimensional setting ($p\gg n$). As such, it might be possible to improve the generalization performance by using a different set of initial parameters better adapted to the high-dimensional setting. This will be investigated in future work. 

However, in this paper, we did not intend to improve the generalization performance by trying to tune the initial parameters for each specific dataset. This was not the point of this project. We rather wanted to highlight our gradient-based methods compare favorably in terms of generalization with benchmark procedures just by using the default initial values in Table \ref{Tab:parameter}.

\begin{table}[http!]
\begin{center}\centering
\begin{tabular}{|l||l|}
\hline
\footnotesize{\textbf{Optimization parameters}}&  \footnotesize{\textbf{Parameter initialization}}\\
\hline
\hline
\begin{tabular}{l|l}
\footnotesize{\textbf{Tolerance}} &  $10^{-4}$\\
\footnotesize{\textbf{Max. iter.}} &  $10^{3}$\\
\footnotesize{\textbf{Learning rate}} &  $0.5$\\
\footnotesize{\textbf{Adam $\beta_1$}} &  $0.5$\\
\footnotesize{\textbf{Adam $\beta_2$}} &  $0.9$\\
\end{tabular} &
\begin{tabular}{l|l}
\footnotesize{\textbf{$T$}} & $30$ \\
\footnotesize{\textbf{$\lambda$}} & $10^{3}$ \\
\footnotesize{\textbf{$\spa$}} & $0_p$ \\
\footnotesize{\textbf{$\kappa$}} & $0.1$ \\
\footnotesize{\textbf{$\mu$}} & $0$ \\
\end{tabular} \\
\hline
\end{tabular}
\end{center}
\caption{Parameters for the \BKKs\, procedures.}
\label{Tab:parameter}
\end{table}

\begin{figure}[http!]
\centering
\includegraphics[scale=0.27]{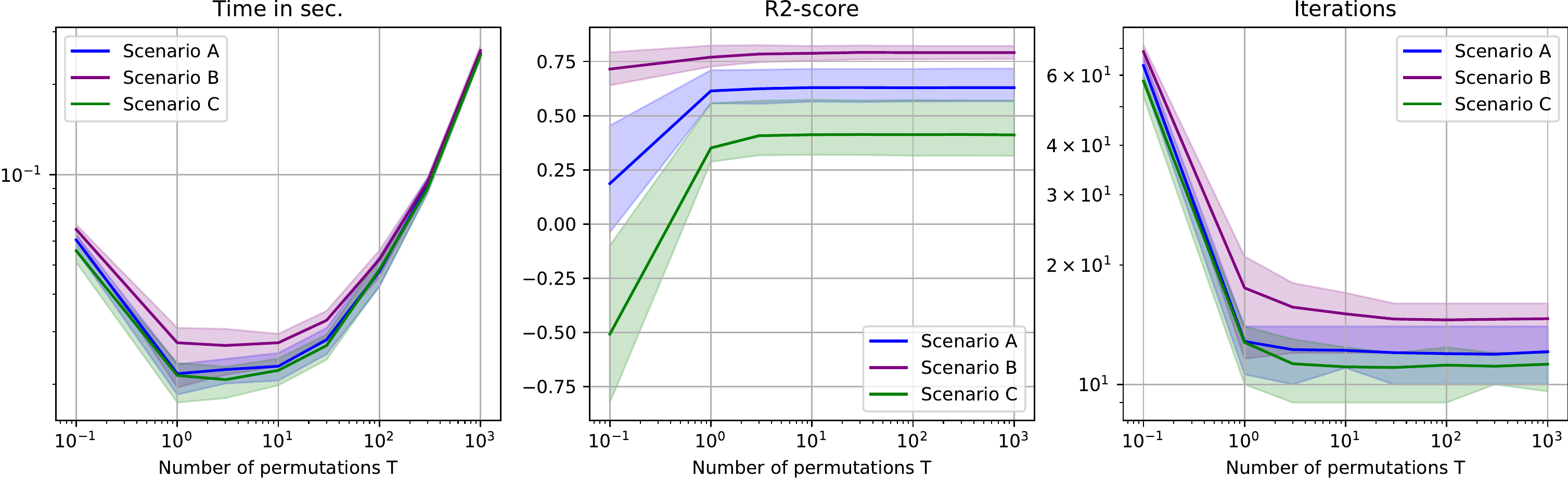}
\caption{\textbf{Synthetic data:} impact of $T$ on the \BKKs\, procedures. } 
\label{fig:syntheticT}
\end{figure}

We also studied the impact of parameter $T$ on the performances of the \BKKs\, procedures on the synthetic data. In Figure~\ref{fig:syntheticT}, the generalization performance ($\mathbf{R^2}$-score) increases significantly from the first added permutation ($T=1$) %as soon as $T=1$
. Starting from $T\approx 10$, the $\mathbf{R^2}$-score has converged to its maximum value. An even more striking phenomenon is the gain observed in the running time when we add $T$ permutations (for $T$ in the range from $1$ to approximately $100$) when compared with the usual empirical risk ($T=0$). Larger values of $T$ are neither judicious nor needed in this approach. In addition, the needed number of iterations for ADAM to converge is divided by $3$ starting from the first added permutation. Furthermore, this number of iterations remained stable (below 20) starting from $T=1$. Based on these observations, \textbf{the hyperparameter T does not require calibration}. We fixed $T=30$ in our experiments even though $T=10$ might have been sufficient.

\paragraph{Performance comparisons.}
%%%%%%%%%%%%%%%%%%%%%%%%%%%%%%%%%%%%%%%%%%

We compare our \BKKs\, procedures against cross-validated Ridge, \LAS and Elastic-net (implemented as RidgeCV,
LassoCV and ElasticnetCV in Scikit-learn \cite{pedregosa2011scikit}) 
on simulated and real datasets. Our procedures are implemented in PyTorch \cite{paszke2019pytorch} on the centered and rescaled response $\bY$.
Complete details and results can be found in the Appendix. In our approach $\theta$ can always be tuned directly on the $train$ set whereas for benchmark procedures like \LAS, Ridge, Elastic net, $\theta$ is typically calibrated on a hold-out $validation$-set using grid-search CV for instance.

\paragraph{Generalisation performance.}
%%%%%%%%%%%%%%%%%%%%%%%%%%%%%%%%%%%%%%%%%%

\begin{figure}[http!]
\centering
 \includegraphics[scale=0.32]{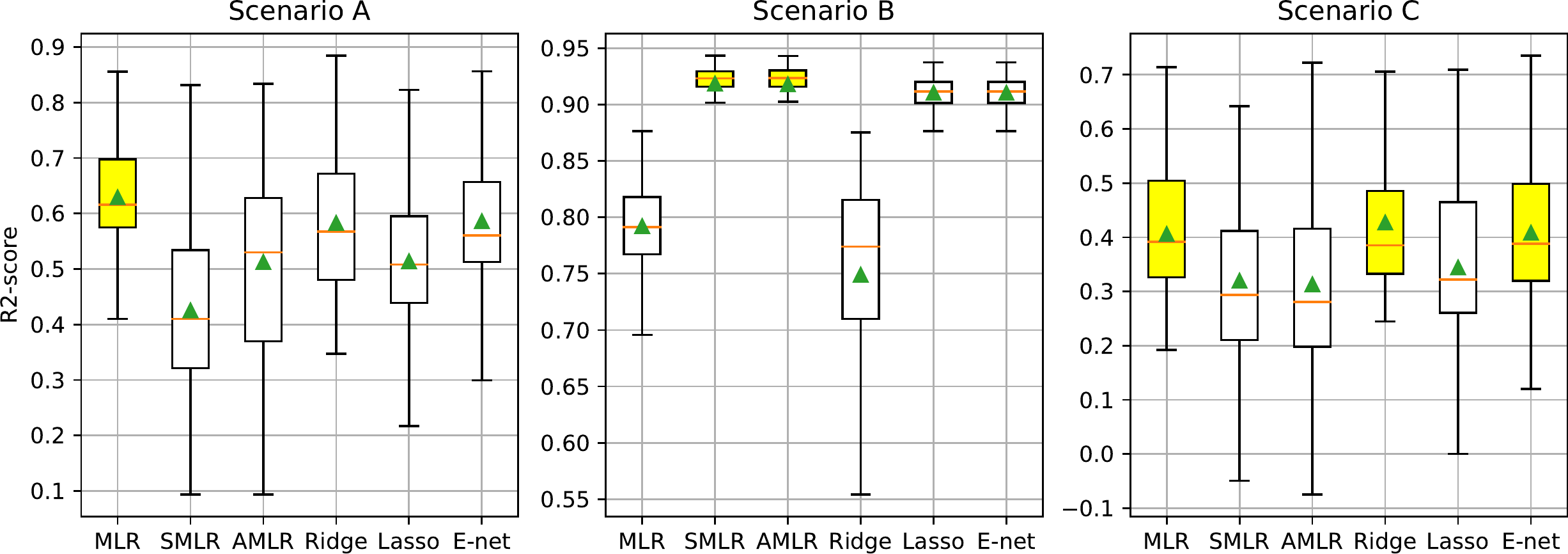}
\caption{\textbf{Synthetic data:} R2-score}
\label{fig:syntheticR2}
 \end{figure}

\begin{figure}[http]
\centering
\includegraphics[scale=0.3]{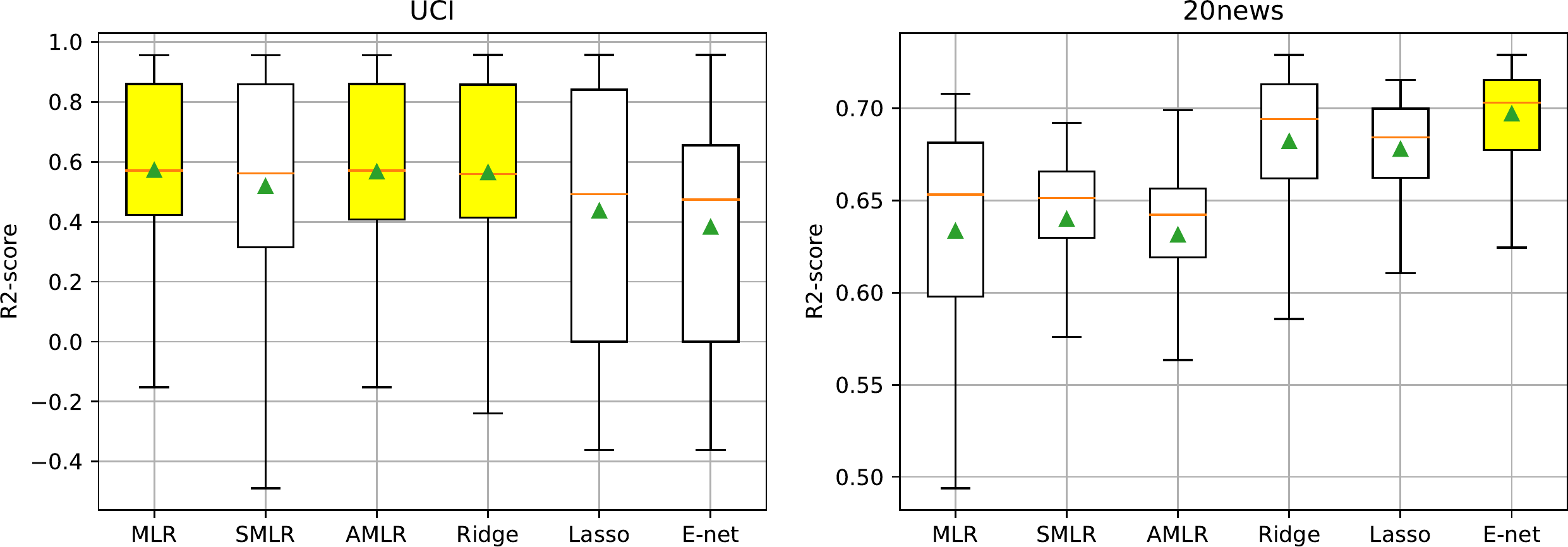}
\caption{\textbf{UCI data} (left) and \textbf{20news data} (right): R2-score}
\label{fig:UCIR2}
\end{figure}

Figures \ref{fig:syntheticR2} and \ref{fig:UCIR2} show the \BKKs\, procedures consistently attain the highest $\mathbf{R^2}$-scores for the synthetic and UCI data according to the Mann-Whitney test over the $M=100$ repetitions. Regarding, the 20news datasets, the \BKKs\, procedures are always within $0.05$ of the best (E-net).


\paragraph{Estimation of $\beta^*$ and support recovery accuracy.}
%%%%%%%%%%%%%%%%%%%%%%%%%%%%%%%%%%%%%%%%%%

\begin{figure}[htp]
\centering
\includegraphics[scale=0.31]{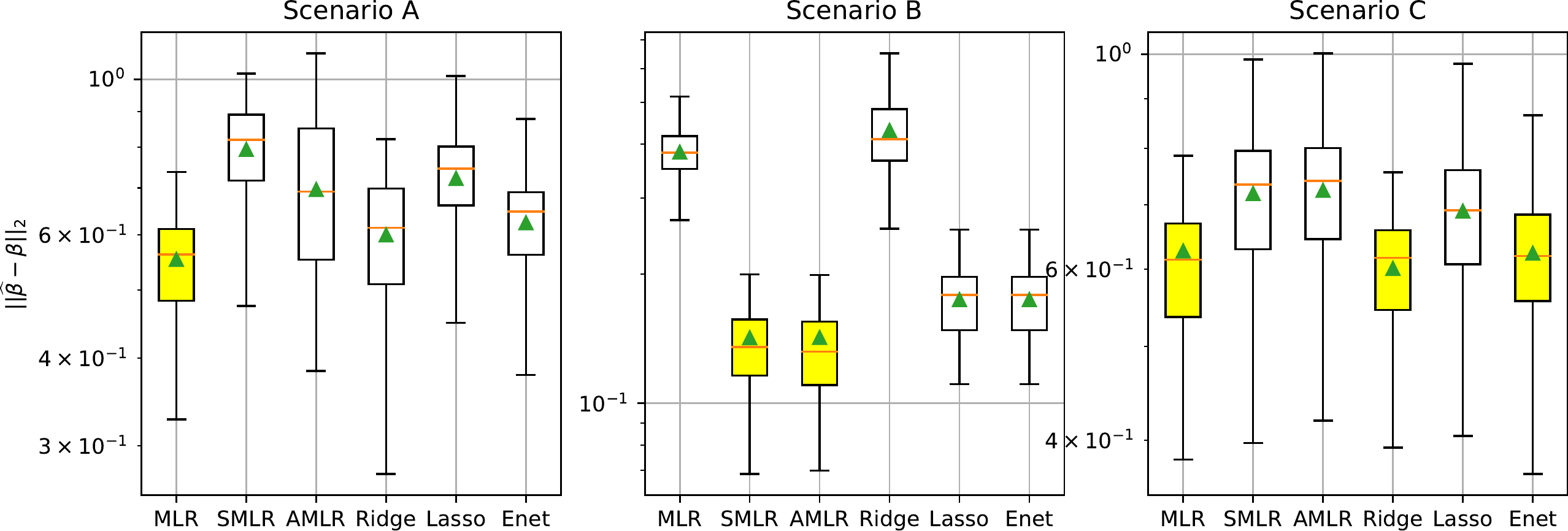}
\caption{ $\beta^*$ estimation  (best in yellow according to MW) }
\label{fig:l2norm}
\end{figure}

For the synthetic data, we also consider the estimation of the regression vector $\beta^*$. We use the $l_2$-norm estimation error $\|\widehat{\beta} - \beta^*\|_{2}$ to compare the procedures. As we can see in Figure~\ref{fig:l2norm}, the \GER\, procedures perform better than the benchmark procedures.

We finally study the support recovery accuracy in the sparse setting (\textbf{Scenario B}). We want to recover the support $J(\be^*) = \left\lbrace j
\, : \, \be^*_j \neq 0 \right\rbrace $.  For our procedures, we build the following estimator $\widehat{J}(\widehat{\be}) = \left\lbrace j
\, : \, |\widehat{\be}_j|>\widehat{\tau} \right\rbrace$ where the threshold $\widehat{\tau}$ corresponds to the first sharp decline of the coefficients $|\widehat{\beta}_j|$. 
Denote by $\#J$ the cardinality of set $J$. The support recovery accuracy is measured as follows: 
\begin{align*}
\label{eq:ACC}
Acc(\widehat{\be}):=\frac{\# \lbrace J(\beta^*) \cap \widehat{J}(\widehat{\be}) \rbrace  + \# \lbrace J^c(\beta^*) \cap \widehat{J}^c(\widehat{\be})\#\rbrace}{p},
\end{align*}

\begin{figure}[http]
\centering
\includegraphics[scale=0.35]{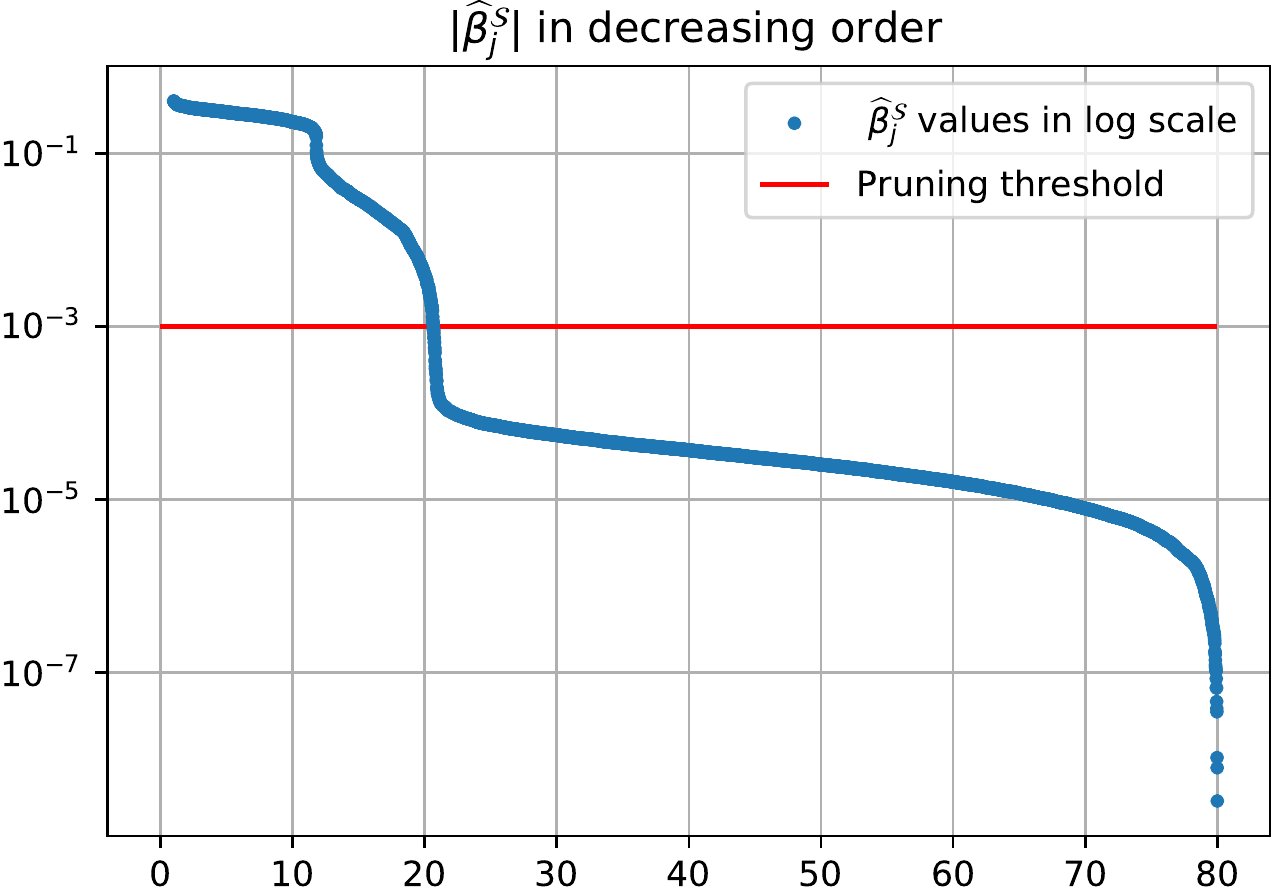}
\caption{Coefficients values (blue) and threshold (red) with $p=80$.  }
\label{fig:pruning}
\end{figure}

Our simulations confirm $\widehat{\be}^{\cS}$ is a quasi-sparse vector. Indeed we observe in Figure \ref{fig:pruning} a sharp decline of the coefficients $|\widehat{\be}^{\cS}_j|$. Thus we set the threshold $\widehat{\tau}$ at $10^{-3}$. 

\begin{figure}[htp]
\centering
\includegraphics[scale=0.35]{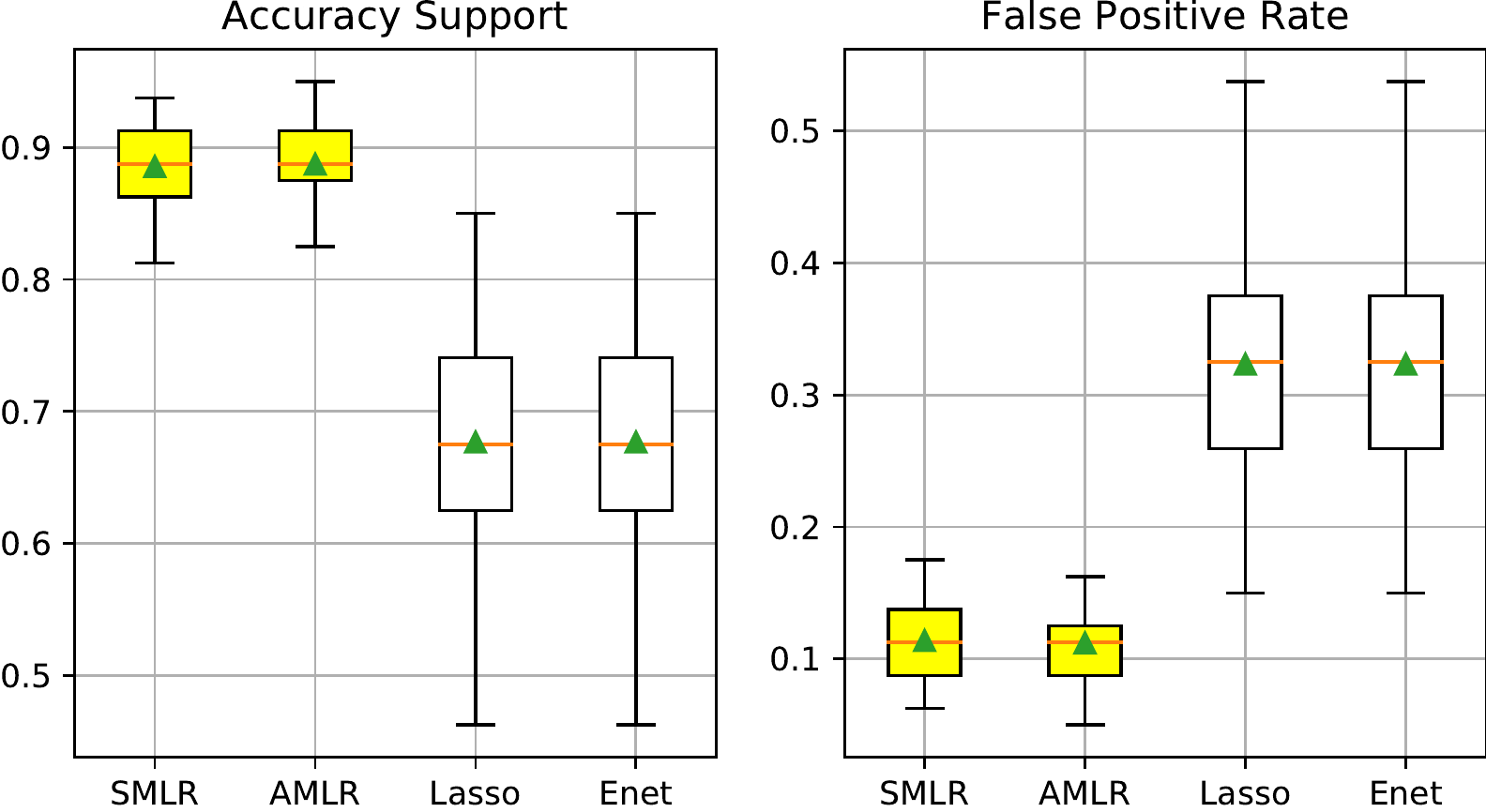}
\caption{Support recovery performance analysis in \textbf{Scenario B} (best in yellow according to MW).}
\label{fig:support}
\end{figure}

Overall, $\widehat{\be}^{\cS}$ and $\widehat{\be}^{\cA}$  perform better for support recovery than the benchmark procedures. Moreover in \textbf{Scenario B} favorable to LASSO, our procedures perform far better (Figure~\ref{fig:support}).

\begin{figure}[htp]
\centering
\includegraphics[scale=0.3]{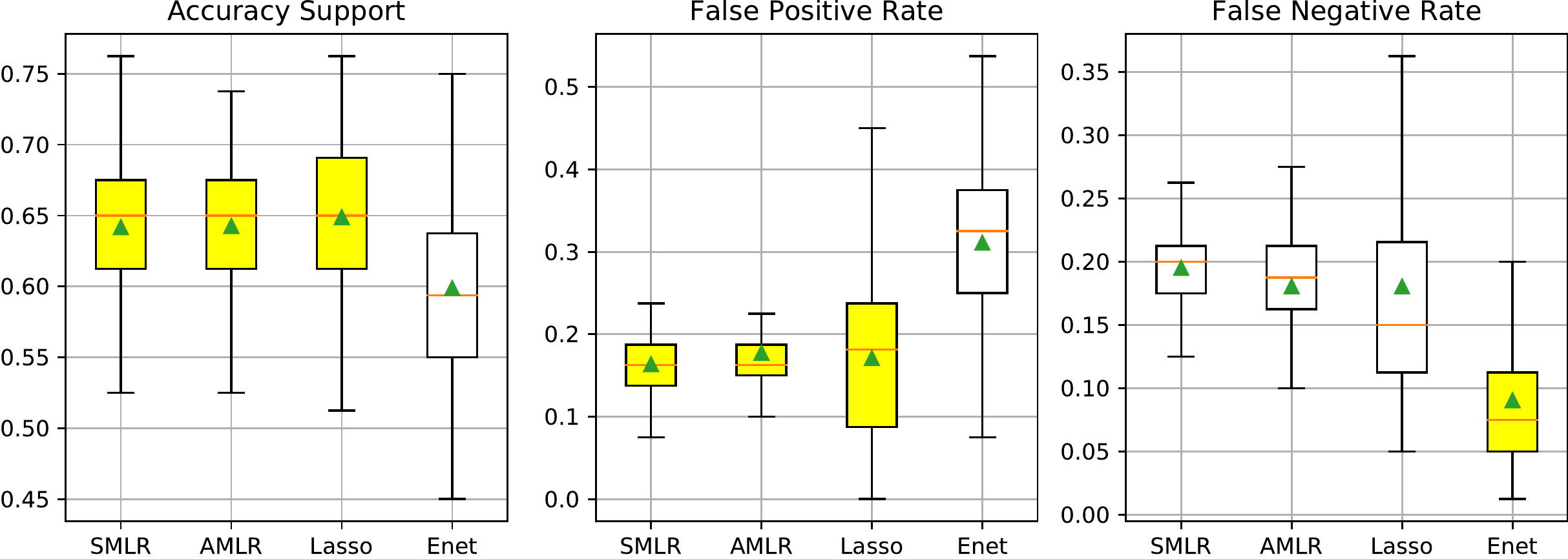}
\caption{Support recovery performance analysis in \textbf{Scenario C} (best in yellow according to MW).}
\label{fig:support2}
\end{figure}

%%%%%%%%%%%%%%%%%%%%%%%%%%%%%%%%%%%%%%%%%%%
\section{Conclusion and future work}
\label{sec:con}
%%%%%%%%%%%%%%%%%%%%%%%%%%%%%%%%%

In this paper, we introduced in the linear regression setting the new \GER\, approach based on a different understanding of generalization. Exploiting this idea, we derived a novel criterion and new procedures which can be implemented directly on the $train$-set without any hold-out $validation$-set. Within \GER, additional structures can be taken into account without any significant increase in the computational complexity.

We highlighted several additional advantageous properties of the \BKK\, approach in our numerical experiments. The \BKKs\, approach is computationally feasible while yielding  statistical performances equivalent or better than the cross-validated benchmarks.
We provided numerical evidence of \GER\, criterion's ability to generalize from the first added permutation. Besides, the strength of our \GER\, procedures stems from their compatibility with gradient-based optimization methods. As such, these procedures can fully benefit from automatic graph-differentiation libraries (such as pytorch \cite{paszke2017automatic} and tensorflow \cite{tensorflow}).

In our numerical experiments, adding more permutations improves the convergence of the ADAM optimizer while preserving generalisation.  As a matter of fact, $T$ does not require any fine-tuning. In that regard, $T$ is not a hyperparameter. Likewise, the other hyperparameters require no tedious initialization in this framework. The same fixed hyperparameters for ADAM and initialization values of the regularization parameters (see Table \ref{Tab:parameter}) were used for all the considered datasets. Noticeably, these experiments were run using high values for learning rate and convergence threshold. Consequently, only a very small number of iterations were needed, even for non-convex criteria ($\GER_{\be^{\cS}}$ and $\GER_{\be^{\cA}}$).

The \GER\, approach offers promising perspectives to address an impediment to the broader use of deep learning. Currently, fine-tuning DNN numerous hyper-parameters often involves heavy computational resources and manual supervision from field experts\cite{smith2018disciplined}. Nonetheless, it is widely accepted that deep neural networks produce state-of-the-art results on most machine learning benchmarks based on large and structured datasets \cite{escalera2018neurips,he2016deep,klambauer2017self,krizhevsky2012imagenet,silver2016mastering,simonyan2014very,szegedy2015going}. By contrast, it is not yet the case for small unstructured datasets, (eg. tabular datasets with less than 1000 observations ) where Random Forest, XGBOOST, MARS, $etc$ are usually acknowledged as the state of the art \cite{shavitt2018regularization}.

These concerns are all the more relevant during the ongoing global health crisis. Reacting early and appropriately to new streams of information became a daily challenge. Specifically, relying on the minimum amount of data to produce informed decisions on a massive scale has become the crux of the matter. In this unprecedented situation, transfer learning and domain knowledge might not be relied on to address these concerns. In that regard, the minimal need for calibration and the reliable convergence behavior of the \GER\, approach are a key milestone in the search for fast reliable regularization methods of deep neural networks, especially in the small sample regime.

Beyond the results provided in this paper, we successfully extended the \GER\, approach to deep neural networks. Neural networks trained with the \BKK\,criterion can reach state of the art results on benchmarks usually dominated by Random Forest and Gradient Boosting techniques. Moreover, these results were obtained while preserving the fast, smooth and reliable convergence behavior displayed in this paper. We also successfully extended our approach to classification on tabular data. All these results are the topics of a future paper which will be posted on Arxiv soon. In an ongoing project, we are also adapting our approach to tackle few-shots learning and adversarial resilience for structured data (images, texts, graphs). We believe we just touched upon the many potential applications of the \GER\, approach in the fields of Machine Learning, Statistics and Econometrics.

\bibliographystyle{plainnat}
\bibliography{BiblioPapier1}

\end{document}